\documentclass[letterpaper, 10 pt, conference]{ieeeconf}  % Comment this line out if you need a4paper

\IEEEoverridecommandlockouts                              % This command is only needed if 
\usepackage[top=19.1mm, bottom=19.1mm, left=19.1mm, right=19.1mm]{geometry}
\usepackage{lipsum}
\usepackage{colortbl}
\usepackage{url}
\usepackage{graphicx}
\usepackage{amsfonts}
\usepackage{soul} % 导入 soul 包
\usepackage{color, xcolor} % 颜色包，color 必须导入，xcolor 建议导入
\usepackage{amsmath}
\usepackage{amssymb}
\usepackage{multirow}
\usepackage{array}
\usepackage[linesnumbered,ruled,vlined]{algorithm2e}
\usepackage{booktabs}
\usepackage[noadjust]{cite}
\usepackage[colorlinks,allcolors=black]{hyperref}
\usepackage{color}
\usepackage[caption=false, font=footnotesize]{subfig}

\usepackage{threeparttable}
\usepackage{arydshln}
\title{\LARGE \bf Enhancing Deep Reinforcement Learning-based Robot Navigation Generalization through Scenario Augmentation}

\author{Shanze Wang$^{1, 2\dagger}$, Mingao Tan$^{2\dagger}$, Zhibo Yang$^{3}$, Xianghui Wang$^{2}$, Xiaoyu Shen$^{2}$, \\ Hailong Huang$^{1}$ and Wei Zhang$^{2}$
\thanks{This work has been submitted to the IEEE for possible publication. Copyright may be transferred without notice, after which this version may no longer be accessible.}
\thanks{*This work is supported by 2035 Key Research and Development Program of Ningbo City under Grant No.2024Z127. (\textit{Corresponding author: Wei Zhang.})}% <-this % stops a space
\thanks{$^{1}$Shanze Wang and Hailong Huang are with the Department of Aeronautical and Aviation Engineering, Hong Kong Polytechnic University. {\tt\small shanze.wang@connect.polyu.hk; hailong.huang@polyu.edu.hk.}}
\thanks{$^{2}$Shanze Wang, Mingao Tan, Xianghui Wang, Xiaoyu Shen and Wei Zhang are with the Ningbo Key Laboratory of Spatial Intelligence and Digital Derivative, Institute of Digital Twin, Eastern Institute of Technology, Ningbo, China. {\tt\small  szwang@eitech.edu.cn; mtan@eitech.edu.cn; xhwang@eitech.edu.cn; xyshen@eitech.edu.cn; zhw@eitech.edu.cn.}}
\thanks{$^{3}$Zhibo Yang is with the Department of Mechanical Engineering, National University of Singapore. {\tt\small  zhibo.yang@u.nus.edu}} 
\thanks{$^{\dagger}$These authors contributed equally to this work as co-first authors.} 
}

\begin{document}
\maketitle
\thispagestyle{empty}
\pagestyle{empty}

\begin{abstract}
This work focuses on enhancing the generalization performance of deep reinforcement learning-based robot navigation in unseen environments.
We present a novel data augmentation approach called scenario augmentation, which enables robots to navigate effectively across diverse settings without altering the training scenario. The method operates by mapping the robot's observation into an imagined space, generating an imagined action based on this transformed observation, and then remapping this action back to the real action executed in simulation. Through scenario augmentation, we conduct extensive comparative experiments to investigate the underlying causes of suboptimal navigation behaviors in unseen environments. Our analysis indicates that limited training scenarios represent the primary factor behind these undesired behaviors. Experimental results confirm that scenario augmentation substantially enhances the generalization capabilities of deep reinforcement learning-based navigation systems. The improved navigation framework demonstrates exceptional performance by producing near-optimal trajectories with significantly reduced navigation time in real-world applications.
% A novel data augmentation method, termed scenario augmentation, is presented, enabling robot to navigate in diverse scenarios without modifying its training scenario. The core idea of scenario augmentation involves mapping the robot's observation into an imagined space, returning the imagined action based on the imagined observation, and remapping the imagined action back to the real action executed in simulation. Through scenario augmentation, extensive comparative experiments are conducted to investigate the causes of undesired navigation behaviors in unseen environments. Analysis using the proposed uncertainty map and local-trajectory map reveals that insufficient training scenarios constitute the primary cause of these undesired behaviors. Experimental results demonstrate that scenario augmentation significantly improves the generalization performance of deep reinforcement learning-based navigation. The enhanced navigation system demonstrates superior performance by generating near-optimal trajectories with substantially reduced navigation time in real-world tasks.
% Supplementary videos of the experiments are available at \href{https://youtu.be/JDgntcdYOKE}{https://youtu.be/JDgntcdYOKE}.
\end{abstract}

\section{Introduction}

Autonomous navigation is important for mobile robots, with SLAM efficiently handling known environments \cite{dissanayake_solution_2001}. However, in unknown or dynamic settings, unreliable maps limit traditional methods, necessitating mapless navigation based on local perception \cite{pfeiffer_reinforced_2018}. Learning-based approaches using deep neural networks \cite{lecun_deep_2015} generate motion commands directly from sensor data, bypassing intermediate processes while handling high-dimensional inputs with computational efficiency even on modest hardware \cite{montero2024dynamic}.
% The capability for autonomous navigation represents a fundamental requirement of mobile robots. Map-based approaches, particularly SLAM, effectively address navigation in known and static scenarios \cite{dissanayake_solution_2001}. However, in unknown and dynamic scenarios, no prior maps are available, and constructed maps are often unreliable. Without a reliable map, the application of classical map-based methods is limited, necessitating robust map-less navigation capabilities using only local perception \cite{pfeiffer_reinforced_2018}.
% In recent years, learning-based approaches have garnered significant attention in the domain of map-less autonomous navigation. Unlike conventional methods, learning-based approaches utilize deep neural networks (DNN) \cite{lecun_deep_2015} as motion controllers, generating motion commands end-to-end. By bypassing time-consuming processes such as map building, state estimation, global path planning, and local motion planning, DNN can directly output control commands from sensor readings. Unlike conventional map-less navigation methods, such as Fuzzy Reactive Control (FRC) \cite{xu_sensor-based_1997}, DRL-based approaches can handle high-dimensional observations like images or dense laser scans. Moreover, the forward propagation of DNN is computationally efficient and can even be executed online by a Raspberry Pi \cite{montero2024dynamic}.

Deep Reinforcement Learning (DRL) \cite{mnih_human-level_2015} leads among learning-based navigation methods, with agents optimizing behavior through environmental interactions and reward signals. DRL surpasses imitation learning in robustness without requiring expert demonstrations \cite{liu_2022_tase}. While physical training presents resource and safety challenges, simulation offers a practical alternative. Advanced simulators with simulation-to-reality techniques \cite{muratore_assessing_2019} enable effective policy transfer to real-world systems, particularly for robots with distance sensors.
\begin{figure}[t]
	\centering
	\includegraphics[width=\linewidth]{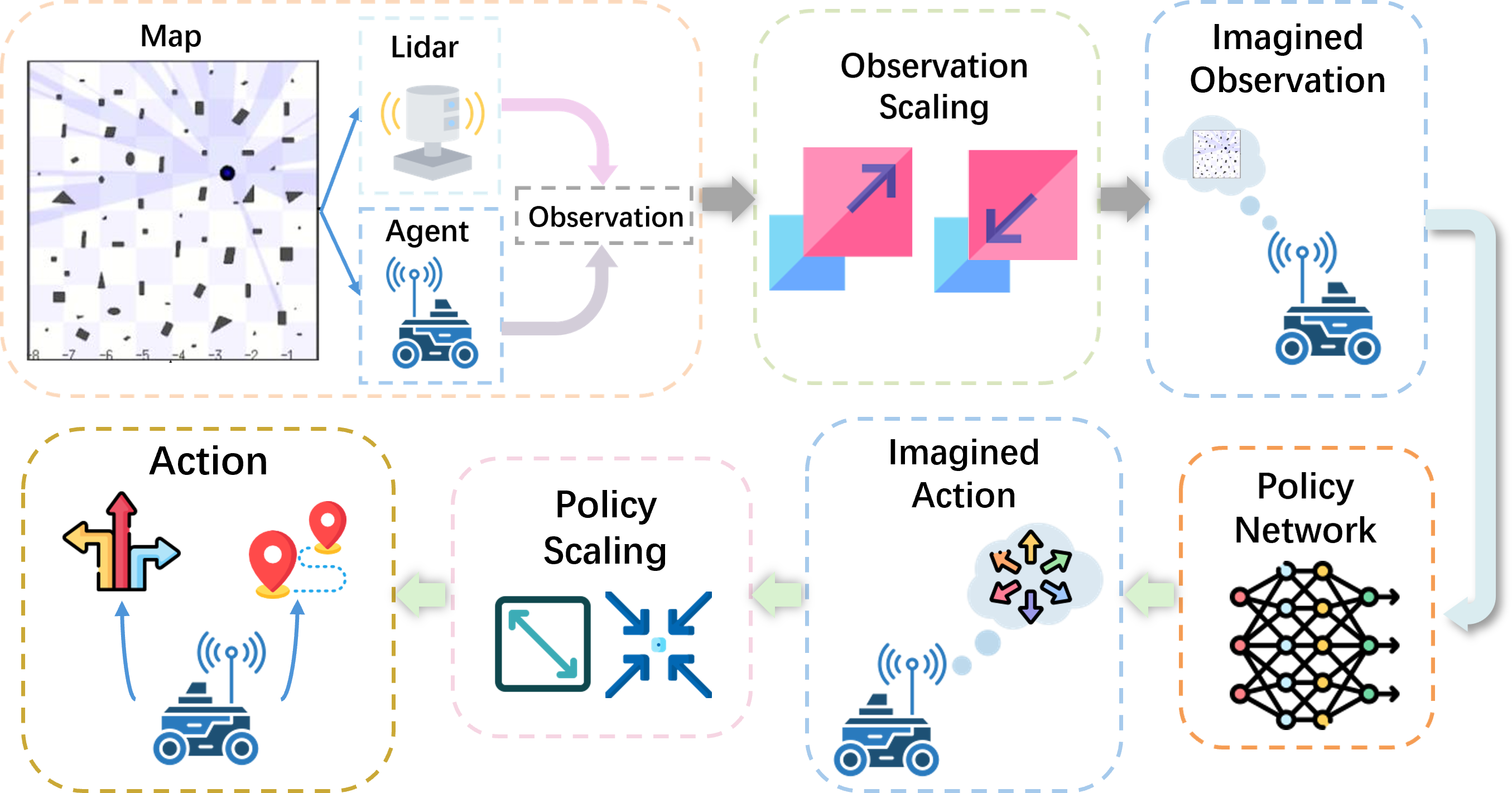}
	\caption{The framework of prorpsed scenario augmentation method.}
	\label{framework}
\end{figure}
Existing research predominantly focuses on optimizing navigation performance within training environments \cite{tai_virtual--real_2017}, \cite{xie_learning_2018}, \cite{xie_learning_2020}, yielding favorable outcomes when test conditions mirror training scenarios \cite{shi_end--end_2019}. However, significant performance degradation occurs in novel environments, where DRL agents exhibit undesired behaviors \cite{tai_virtual--real_2017, xie_learning_2020}.
Current DRL navigation approaches typically employ a severely constrained set of training environments \cite{zhu_target-driven_2017}, with most studies limiting training to either one \cite{shi_end--end_2019} or two \cite{xie_learning_2018, xie_learning_2020} carefully designed scenarios. While enhancing environmental diversity improves generalization capabilities \cite{chen2024semantically}, it simultaneously increases computational demands, highlighting the need for resource-efficient generalization methods \cite{hirose_exaug_2023}. This generalization challenge \cite{cobbe_quantifying_2019} arises when real-world observations diverge substantially from training data, compromising the agent's ability to transfer learned behaviors to new situations.

This paper presents scenario augmentation, a computationally efficient data augmentation method that generates diverse training scenarios without modifying the physical environment layout. Through systematic analysis, we identify insufficient training scenario diversity as the primary factor in performance degradation. Experimental results confirm that scenario augmentation significantly enhances the generalization capability of DRL navigation agents. The main contributions of this paper are:
\begin{itemize}
\item A novel scenario augmentation method that improves DRL navigation agents' generalization by creating dynamic mappings between real and imagined spaces, enabling diverse training experiences while maintaining the same physical environment.
\item Comprehensive experimental validation across multiple simulated and real-world environments, demonstrating the method's effectiveness and robustness in both complex static settings and dynamic pedestrian scenarios.
\end{itemize}

\section{Related Works}

Deep Reinforcement Learning (DRL) enables adaptive learning in dynamic environments \cite{Wang_2024_tase} and optimizes complex, high-dimensional policies, thus enhancing navigation efficiency and robustness. Although algorithms such as DQN \cite{van2016deep}, PPO \cite{schulman2017proximal}, and SAC \cite{haarnoja2018soft} have been widely implemented, they are not specifically designed for mapless navigation tasks \cite{gao_2024_tase}. Consequently, they may encounter challenges such as sparse rewards and limited generalization in complex, crowded environments.

The capacity for generalization in DRL becomes paramount when testing environments differ from training environments. Navigation agents are prone to overfitting, which may result in suboptimal decisions in unfamiliar settings \cite{kalidas_deep_2023}. Various techniques have been explored to improve DRL generalization, with research predominantly focusing on addressing overfitting to restricted observations and adapting to new environments \cite{kirk_survey_2023}.

Researchers have proposed various approaches to navigation challenges. Yang et al. \cite{10271559} enhanced safe navigation in crowded environments using deep reinforcement learning with a local risk map that captured human interactions. Reinis et al. \cite{9645287} developed a system that selected optimal waypoints and integrated learned motion policies for navigation without prior maps in both static and dynamic environments. Xie et al. \cite{10089196} created a control strategy combining LiDAR data, pedestrian kinematics, and subgoal points to generate steering and velocity commands.

Studies have shown that neural networks favor simple functions, potentially limiting their effectiveness in reinforcement learning tasks \cite{fort_deep_2020}. The bias toward smooth functions can negatively impact value approximation in DRL \cite{valle-perez_deep_2019}. Yang et al. \cite{yang_overcoming_2022} and Brellmann et al. \cite{brellmann_fourier_2023} proposed adjusting Fourier feature scales to capture high-frequency components, though this approach risks fitting noise and reducing generalization capability with limited training data.

Imitation learning accelerates DRL training \cite{pfeiffer_reinforced_2018} but requires significant design effort. Shi et al. \cite{shi_end--end_2019} and Cai et al. \cite{cai_2024_tase} improved training efficiency using intrinsic and unrelated rewards, though generalization to novel scenarios remains uncertain. Raileanu and Fergus \cite{raileanu_decoupling_2021} found that decoupling value approximation and policy networks improves generalization, despite increased computational demands and potential training instability.

\section{Problem Formulation}

\begin{figure}[t]
	\centering
	\includegraphics[width=0.45\linewidth]{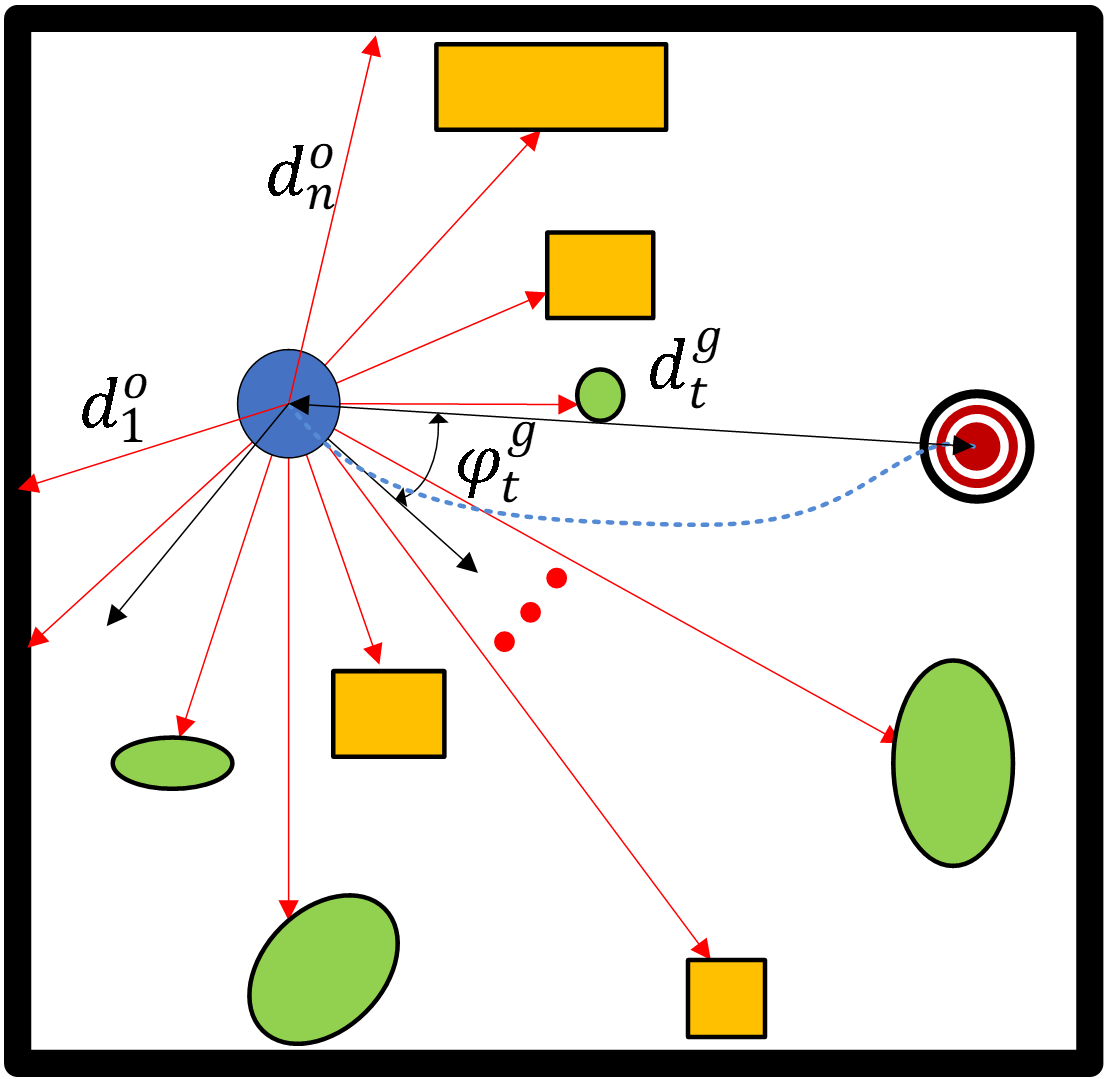}
	\caption{Illustration of the goal-driven robot navigation problem.}
	\label{fig1}
\end{figure}

Goal-driven robot navigation can be discretized into a sequential decision-making process. As illustrated in Fig. \ref{fig1}, the robot's task is to reach its designated goal while successfully navigating around obstacles. The goal's relative position $s_t^g=\left\{d_t^g,\varphi_t^g\right\}$ is determined using either pre-existing maps or localization sensors, including WIFI and microphone arrays. The system employs a DNN controller parameterized by $\theta$, which implements a control policy $\pi_\theta$. The controller accepts input $s_t=\left\{s_t^g,s_t^o\right\}$, where $s_t^o$ denotes local environmental data acquired through 2D Lidar scans. The Lidar sensor's coordinate frame coincides with the robot's local reference frame, centered between the drive wheels. At time step $t$, the policy $\pi_\theta$ generates velocities $a_t=\left\{v_t,\omega_t\right\}$ based on input $s_t$, resulting in a reward $r_t$. The objective is to derive the optimal policy $\pi_\theta^\ast$ that maximizes the expected cumulative reward $G_t=\Sigma_{\tau=t}^T\gamma^{\tau-t}r_\tau$, where $\gamma$ denotes the discount factor. The optimal policy $\pi_\theta^\ast$ must demonstrate robust performance across all possible scenarios, extending beyond the training environment.

\section{Method}

The proposed navigation framework combines a deep reinforcement learning backbone with scenario augmentation to achieve robust generalization. The framework of scenario augmentation is shown in Fig. \ref{framework}. During navigation, when the agent receives an observation in simulation, the observation scaling module maps it to an imagined observation, which is then processed by the policy network to generate an imagined action. This imagined action is subsequently transformed by policy scaling back into an executable action in the simulation environment. This bidirectional mapping between real and imagined spaces enables the agent to learn from diverse scenarios while operating in a single physical environment.
% Through observation scaling and policy scaling, the framework enables the agent to learn from diverse scenarios without requiring additional physical environments, while maintaining computational efficiency. This integrated approach significantly enhances the agent's ability to navigate in previously unseen environments.

\subsection{Learning Navigation with SAC}
Once the navigation problem is formulated as a decision-making process, DRL algorithms can be directly applied to identify the optimal policy. The DRL algorithm employed in this paper is the delayed soft actor critic (SAC) \cite{haarnoja_soft_2018}, which incorporates the delayed policy updating technique from twin delayed DDPG (TD3) \cite{fujimoto_addressing_2018} to stabilize the training process of SAC.

\subsubsection{Neural network architecture}
\begin{figure}[t]
	\centering
	\includegraphics[width=\linewidth]{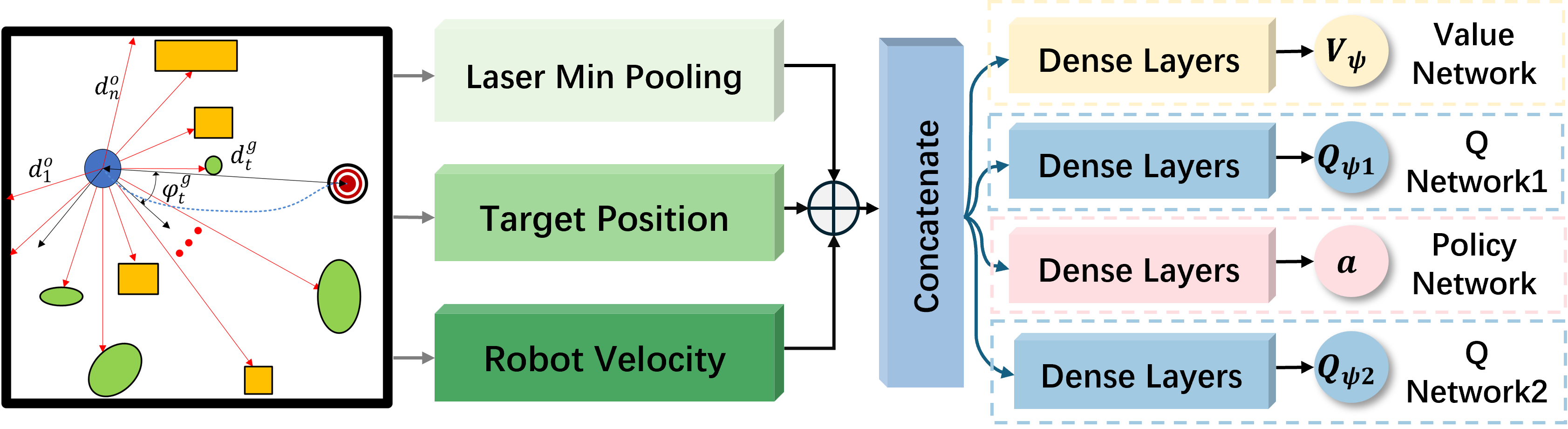}
	\caption{SAC Neural network architecture.}
	\label{fig2}
\end{figure}

The neural network architecture for delayed SAC is presented in Fig. \ref{fig2}, wherein the policy network, value network, and two Q networks are parameterized by $\theta,\psi,\ \phi_1,\phi_2$, respectively. Each network consists of three hidden dense layers, which exhibit superior generalization capabilities compared to convolutional layers \cite{pfeiffer_reinforced_2018}. To optimize computational efficiency, the original laser scan data (1080 laser beams) undergoes 1D minimum down-sampling, reducing the dimensionality to $N$ values ($N=30$). The computation of each down-sampled value $o_{\min}^i$ is defined as:
\begin{equation}
\begin{aligned}
d^o_{\mathrm{min},i} = \min(d^o_{i\cdot k+1}, d^o_{i\cdot k+2}, \cdots, d^o_{i\cdot k+k-1})
\end{aligned}
\end{equation}
where $i$ denotes the index of the down-sampled laser scans, and $k$ represents the down-sampling window length ($k=36$). The state vector $s$, comprising down-sampled laser scans, relative goal position, and robot velocities, is processed by the policy network and value network to compute action $a=\pi_\theta(s)$ and value function $V_\psi(s)$, respectively. The two Q networks take $s$ and $a$ as input and return Q values $Q_{\phi_1}\left(s,a\right)$ and $Q_{\phi_2}(s,a)$. This implementation employs a squashed Gaussian SAC policy, constraining the sampled actions to the interval $[-1,1]$. During the training process, the sampled action $a_{sam}(s|\theta)$ from the squashed Gaussian policy is computed as:
\begin{equation}
\begin{aligned}
a_{sam}(s|\theta)=\tanh{\left(\mu_\theta(s)+\sigma_\theta(s)\odot\zeta\right)\ }
\end{aligned}
\end{equation}
where $\mu_\theta\left(s\right)$ and $\sigma_\theta(s)$ denote the mean and standard deviation of the Gaussian policy, and $\zeta \sim \mathcal{N}(0,1)$.

\subsubsection{Reward function}
% To model the objective of goal-reaching and obstacle-avoidance, the reward function of robot navigation should contain a positive part, $r_{\text{reach}}$, for encouraging goal-reaching and a negative part, $r_{\text{crash}}$, for punishing collision. As the above rewards are too sparse to be received during training, similar to \cite{pfeiffer_reinforced_2018, tai_virtual--real_2017, xie_learning_2018}, an additional dense part is adopted as:
The reward function for robot navigation incorporates two primary components: a positive reward $r_{\text{reach}}$ incentivizing goal-reaching behavior and a negative reward $r_{\text{crash}}$ penalizing collisions. Given that these reward signals occur sparsely during training, we implement an additional dense reward component, following approaches in \cite{pfeiffer_reinforced_2018, tai_virtual--real_2017, xie_learning_2018}, defined as:
\begin{equation}
\begin{aligned}
r_t^{\text{dense}}=\left\{\begin{matrix}r_{\text{reach}},&\text{if reaches the goal,}\\r_{\text{crash}},&\text{if collides,}\\c_1\left(d_t^g-d_{t+1}^g\right),&\text{otherwise.}\\\end{matrix}\right.
\end{aligned}
\end{equation}
where $c_1$ represents a scaling coefficient. Subsequently, the reward function proposed in \cite{xie_learning_2018, zhang_learn_2020} is designated as the dense reward. The dense reward function facilitates training convergence by incentivizing progressive movement toward the target through immediate reward optimization. Nevertheless, this dense component may contradict the time-optimal trajectory objective, potentially inducing suboptimal policy convergence \cite{xie_learning_2018}. The elimination of the dense component yields a sparse reward function that only provides feedback at the terminal states, as expressed in Equation (\ref{eq4}). 
\begin{equation}
\begin{aligned}
r_t^{\text{sparse}}=\left\{\begin{matrix}r_{\text{reach}},&\text{if reaches the goal,}\\r_{\text{crash}},&\text{if collides.}\\\end{matrix}\right.
\label{eq4}
\end{aligned}
\end{equation}
% where $c_1$ is a scaling constant and $c_2$ is a positive constant for punishing staying still. In the rest of this paper, the reward function in reference \cite{xie_learning_2018, zhang_learn_2020} is referred to as the dense reward. The dense reward function can reduce training difficulty by driving the robot to move closer to its target for a higher immediate reward. However, the dense part sometimes conflicts with the shortest-time requirement, leading the agent to learn a suboptimal policy \cite{xie_learning_2018}. Replacing the dense part with a small negative constant $c_3$, i.e., a constant time penalty, makes the navigation objective unbiased, and we refer to this reward function shown in Equation (\ref{eq4}) as the sparse reward function. 
The sparse reward function poses significant challenges in learning navigation policies, as positive rewards are exclusively obtained upon goal achievement. Subsequently, we examine the impact of both reward formulations on the agent's generalization capabilities.

\subsection{Scenario Augmentation}
\begin{figure}[t]
    \centering
    
        \subfloat[]{
        \includegraphics[width=0.48\linewidth]{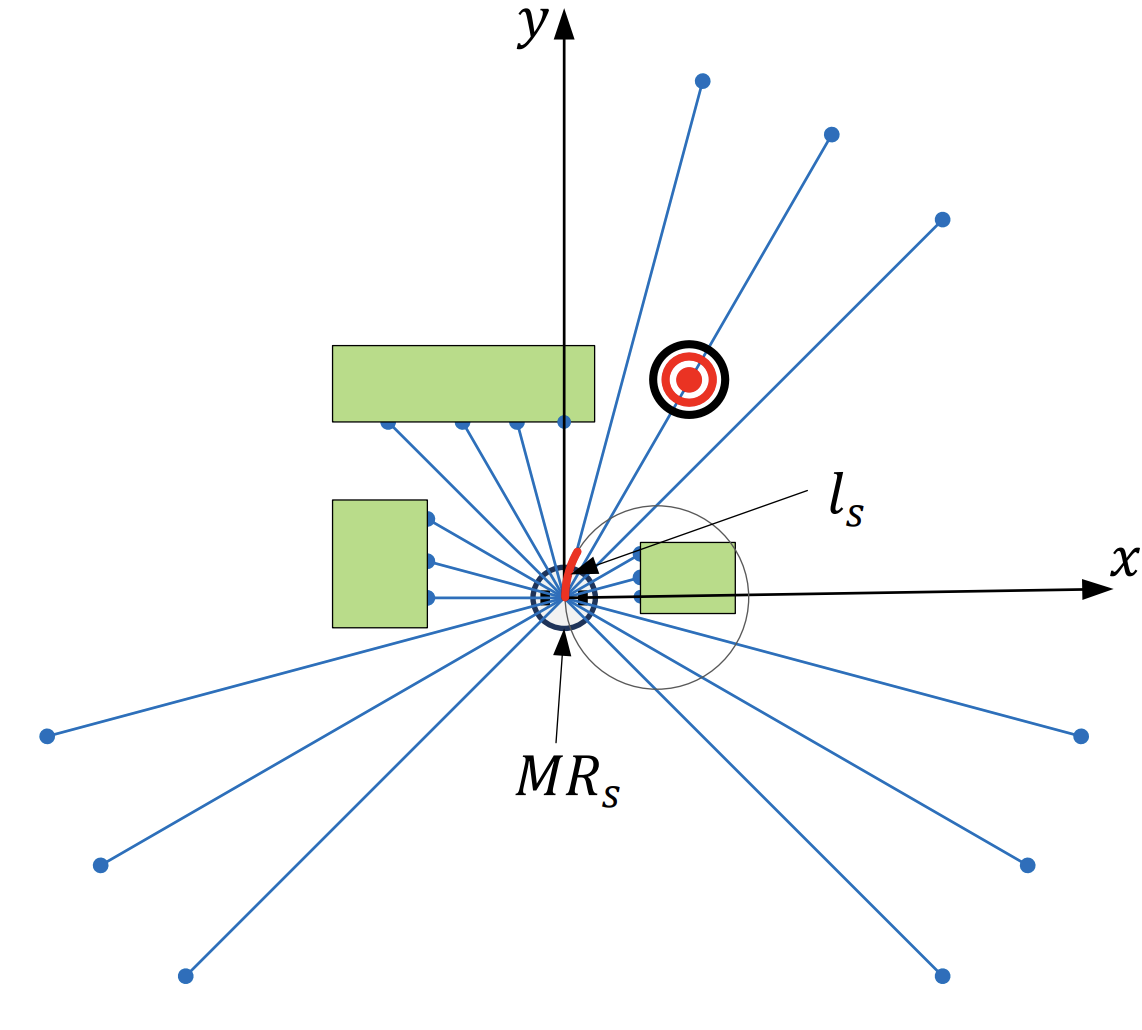}\label{fig3-1}}
        \subfloat[]{
        \includegraphics[width=0.48\linewidth]{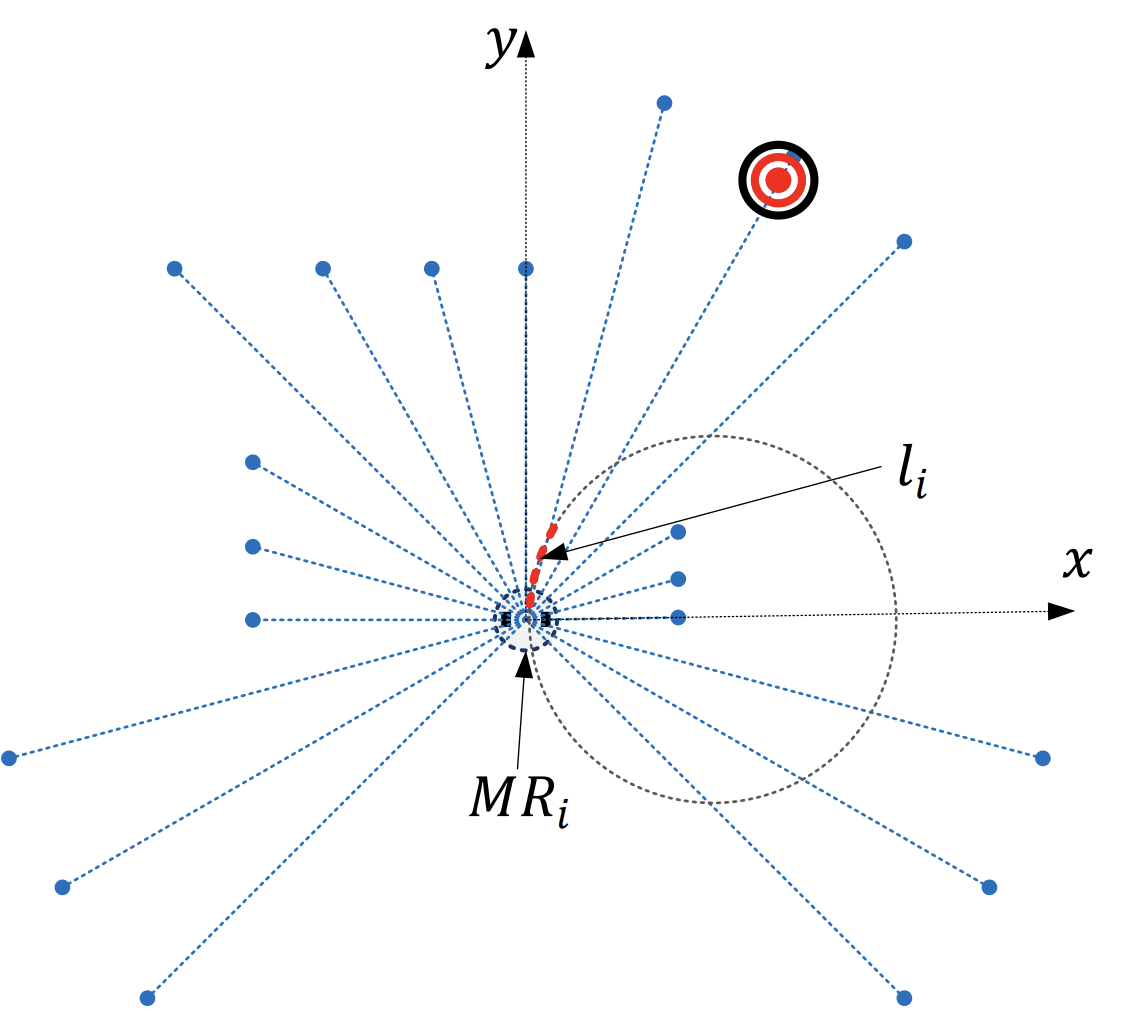}\label{fig3-2}}
        \caption{Overview of scenario augmentation. (a) Robot in simulated space. (b) Robot in imagined space.}
        \label{fig3}
\end{figure}

% Within most robot simulators, including Stage\_ROS, Gazebo, and V-REP \cite{RobotRobotics}, the scenario layout is fixed once the simulated world is loaded. For training navigation agents in multiple scenarios, existing studies predominantly follow a curriculum-learning approach, pre-training the agent in a simple scenario before retraining it in more complex ones. Nevertheless, designing training scenarios for robot navigation requires substantial resources, resulting in a limited number of scenarios. According to Goodfellow et al., enhancing a machine learning model's generalization capability is most effectively achieved through increased training data \cite{goodfellow2016deep}. To efficiently create diverse scenarios for laser-based robot navigation tasks in simulation, we propose scenario augmentation, which generates an infinite number of scenarios without modifying the physical layout of the training scenario. The proposed framework, illustrated in Fig. \ref{fig3}, comprises two stages: 1) observation scaling, where the robot's observations in simulation are amplified into a scaled form (imagined observations); 2) policy scaling, where the policy network returns imagined velocities, which are scaled back to the robot's real velocities in simulation. The mobile robot in simulation is referred to as $MR_s$, and the robot in the imagined scenario as $MR_i$. The two steps of scenario augmentation are detailed below.
Conventional robot simulators, such as Stage\_ROS, Gazebo, and V-REP \cite{RobotRobotics}, maintain a static scenario layout after initial world loading. Existing methodologies for multi-scenario navigation training primarily utilize a curriculum-learning paradigm, wherein agents are initially trained in simplified environments before progressing to more complex scenarios. However, the development of diverse training scenarios demands considerable resources, constraining the available scenario quantity. Goodfellow et al. established that model generalization capability correlates directly with training data diversity \cite{goodfellow2016deep}. We propose scenario augmentation as an efficient methodology for generating diverse laser-based navigation environments, enabling infinite scenario variations while preserving the original physical layout. The proposed framework (Fig. \ref{fig3}) encompasses two primary stages: 1) observation scaling, which transforms simulated robot observations into scaled representations (imagined observations), and 2) policy scaling, which converts network-generated imagined velocities back to realistic robot velocities. We denote the simulated mobile robot as $MR_s$ and its imagined counterpart as $MR_i$. The implementation of these augmentation stages is elaborated as follows.

\subsubsection{Observation-Scaling}
The observation-scaling process transforms current observations into their corresponding imagined representations. Fig. \ref{fig3-1} illustrates $MR_s$ executing a goal-reaching navigation task in the simulation environment. The observation is defined as $s_s=\left\{s_s^{dd},s_s^{di}\right\}$, where $s_s^{dd}=\left\{o_s,v_s,d_s^g\right\}$ denotes distance-dependent observations, and $s_s^{di}=\left\{\omega_s,\varphi_s^g\right\}$ denotes the distance-independent observations. 
For each episode initialization, we introduce a scaling factor $\rho$, defined as the ratio of the dimension of the imagined space to the dimension of the simulated space, which is sampled and applied to the distance-dependent observation $s_s^{dd}$. Specifically, all the distance-dependent observations are amplified by a factor of $\rho$, while the distance-independent observations remain unchanged. Accordingly, the resulting imagined observations $s_i$ are as follows:
\begin{equation}
\begin{aligned}
s_i=\left\{\min{\left\{\rho s_s^{dd},\mathcal{T}(s_s^{dd})\right\}},s_s^{di}\right\}
\label{eq10}
\end{aligned}
\end{equation}
where $\mathcal{T}(\cdot)$ is a threshold function, and $\mathcal{T}(s_s^{dd})$ returns the upper limits of all the elements in $s_s^{dd}$. The imagined observation, as shown in Fig. \ref{fig3-2}, is different from the current observation and cannot be seen in the original training scenario.

\begin{algorithm}[t]

\SetAlgoLined
 Initialize policy parameters $\theta$, Q-value function parameters $\phi_1, \phi_2$, value function parameter $\psi$, empty replay buffer $\mathcal{B}$ \;
 \While {not converge}
 {  Initialize step counter $T$\;
     Sample scenario scaling factor $\rho$ with Eq. (12)\;   
 \While {not terminate}
   {
      Obtain observation $s_s$ in simulation, calculate imagined observation $s_i$ with Eq. (10)\;
      Sample imagined velocities $a_i \sim \pi_\theta$\;
      Calculate the velocities $a_s$ with Eq. (11) and execute it in simulation\;
      Obtain next observation $s'_s$\;
      Calculate $s'_i$, reward $r$ and the terminal signal $d_i$\;
      Store $\{s_i, a_i, r, s'_i, d_i\}$ in replay buffer $\mathcal{B}$\;
      $T \gets T + 1$}
    \For {$t = 1$ to $T$}
{        Sample a minibatch from replay buffer $\mathcal{B}$\; 
        Update $\phi_1, \phi_2, \psi$\; 
        \If {$t \mod 2 = 0$}
            {Update $\theta$\;}}
      }
\caption{Learning robot navigation with Scenario Augmentation}
\label{al1}
\end{algorithm}

\subsubsection{Policy-Scaling}
The policy-scaling process determines the simulated robot's action vector $a_s$. Given imagined observation $s_i$ as input, the policy network returns the imagined velocities $a_i=\pi_\theta\left(s_i\right)=\left\{v_\theta\left(s_i\right),\omega_\theta\left(s_i\right)\right\}$. Following the dynamic window approach (DWA) \cite{fox_dynamic_1997}, the robot is assumed to move with constant velocities within a control cycle, the generated one-step trajectory $l_i$ of $MR_i$ is a circular arc (shown in Fig. \ref{fig3-2}). 
As the dimension of imagined space is $\rho$ times that of the simulated space, the length of the simulated robot's trajectory $l_s$ should be the same as the length of $l_i$ after being amplified by a factor of\ $\rho$, while maintaining identical radians between both trajectories. Accordingly, contrary to the observation-scaling process, the distance-dependent linear velocity is compressed by a factor of $\rho$, while the angular velocity remains unchanged. As a result, the velocity $a_s$ should be:
\begin{equation}
\begin{aligned}
a_s = \left(\frac{v_\theta(s_i)}{\rho}, \omega_\theta(s_i)\right)
\label{eq11}
\end{aligned}
\end{equation}

Through the integration of observation-scaling and policy-scaling mechanisms, the navigation agent accesses observations beyond those available in the original simulation environment. The continuous nature of scaling factor $\rho$ enables the generation of theoretically infinite unique imagined scenarios.
% With observation-scaling and policy-scaling, the navigation agent can receive observations that cannot be encountered in its simulated scenario. Since the scaling factor $\rho$ is continuous, the number of imagined scenarios that could be created is theoretically infinite.

\begin{figure}[t]
	\centering
	\includegraphics[width=0.65\linewidth]{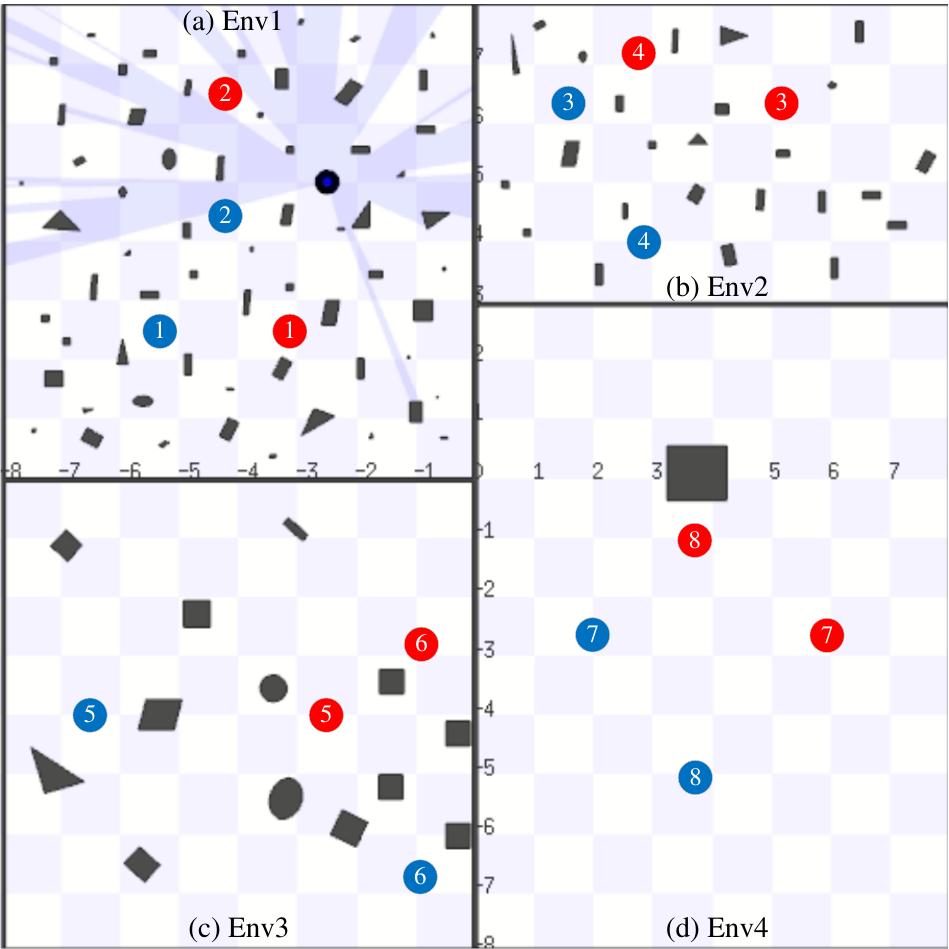}
	\caption{Simulation environments. (a) Training environment (Env1). (b)-(d) Testing environments (Env2-4).}
        \label{fig4}
\end{figure}

\subsection{Learning Navigation with Scenario Augmentation}
The proposed scenario augmentation can be easily incorporated into the learning framework for robot navigation. The pseudo-code of learning robot navigation with scenario augmentation is provided in Algorithm \ref {al1}. At the beginning of each episode, the scenario scaling factor is sampled as follows:
\begin{equation}
\begin{aligned}
\rho=\left\{\begin{matrix}1.0,&\epsilon \geq P\\U\left[1.0,M\right],&\epsilon<P\\\end{matrix}\right.
\end{aligned}
\end{equation}
where $\epsilon \sim U\left[0,1\right]$, $P$ denotes the scenario augmentation probability threshold, and $M$ represents the scaling factor's upper bound. This formulation enables navigation in both simulated and imagined scenarios, while the scaling factor's upper bound $M$ regulates the probability of empty scenario exploration. 
During each episode, simulation observations $s_r$ are transformed into imagined observations $s_i$ via the scaling factor according to Eq. (\ref{eq10}).
The DNN controller generates imagined action $a_i$ based on policy $\pi_\theta$, which is then transformed to real action $a_r$ through Eq. (\ref{eq11}) for simulation execution. 
Following state transition, $MR_r$ obtains subsequent observation $s_r^\prime$, which is transformed into imagined observation $s_i^\prime$. The system computes reward $r_i$ and terminal signal $d_i$ for the imagined transition, storing the transition tuple $\left\{s_i,a_i,r_i,s_i^\prime,d_i\right\}$ in replay buffer $\mathcal{B}$. 
Episode termination occurs upon destination achievement, collision detection, or reaching maximum step count. The system then performs minibatch training to update all trainable parameters. Network updates are executed $T$ times for value and Q networks (where $T$ denotes total episode steps), while the policy network undergoes $T/2$ updates utilizing delayed update optimization.

\section{Experiments} \label{exp}
\subsection{Model training}
To evaluate the efficacy of the scenario augmentation method, we conducted systematic comparisons with multiple baseline approaches:

\begin{itemize}
    \item DDPG\_SG: DDPG with Stochastic Guidance \cite{xie_learning_2020} is an advanced method that uses a Proportional-Integral-Derivative (PID) controller and an Obstacle Avoidance (OA) controller, especially in the early stages of training. While DDPG\_SG learns from these controllers over time to improve performance, it still struggles to generate optimal straight-line navigation strategies in open environments, even after training with the PID controller. Instead, the model develops suboptimal behaviors, indicating that training with expert controllers does not always result in transferring their optimal characteristics.
    
    \item SAC with Different Reward Functions: We implemented two variants:
    \begin{itemize}
        \item SAC\_DR: SAC with Dense Rewards;
        \item SAC\_SR: SAC with Sparse Rewards.
    \end{itemize}
    This comparison examines if performance degradation is caused by the reward function. Dense rewards can speed up training by offering immediate feedback for goal-directed movement, but may lead to suboptimal trajectory learning \cite{xie_learning_2018}. In contrast, sparse rewards give feedback only at terminal states, potentially leading to better policy development.

    \item SAC\_PreT: SAC with Pre-trained Networks. This method was trained in Env4 to see if continuous training in similar scenarios is needed to maintain strong performance, even with pre-training. It suggests that while curriculum learning can speed up initial training, it may not guarantee lasting optimal performance without continuous exposure to the environment.
    
    \item SAC\_SAug: SAC with Scenario Augmentation (our proposed method). This approach addresses the limitations observed in the above baselines by dynamically adapting to different scenarios while maintaining consistent performance across environments.
\end{itemize}

Further experimental analysis included the implementation of SAC\_SR\_v2 and SAC\_SAug\_v2, developed as Dropout-free versions of SAC\_SR and SAC\_SAug, respectively. Unlike previous studies that evaluate DRL-agents within a single scenario, this study employs Env1 (Fig. \ref{fig4}a) for training and conducts performance evaluation across four distinct scenarios (Env1-4, Fig. \ref{fig4}a-d). SAC\_PreT utilizes Env4 as the pre-training scenario.

\begin{figure}[t]
	\centering
	\includegraphics[width=\linewidth]{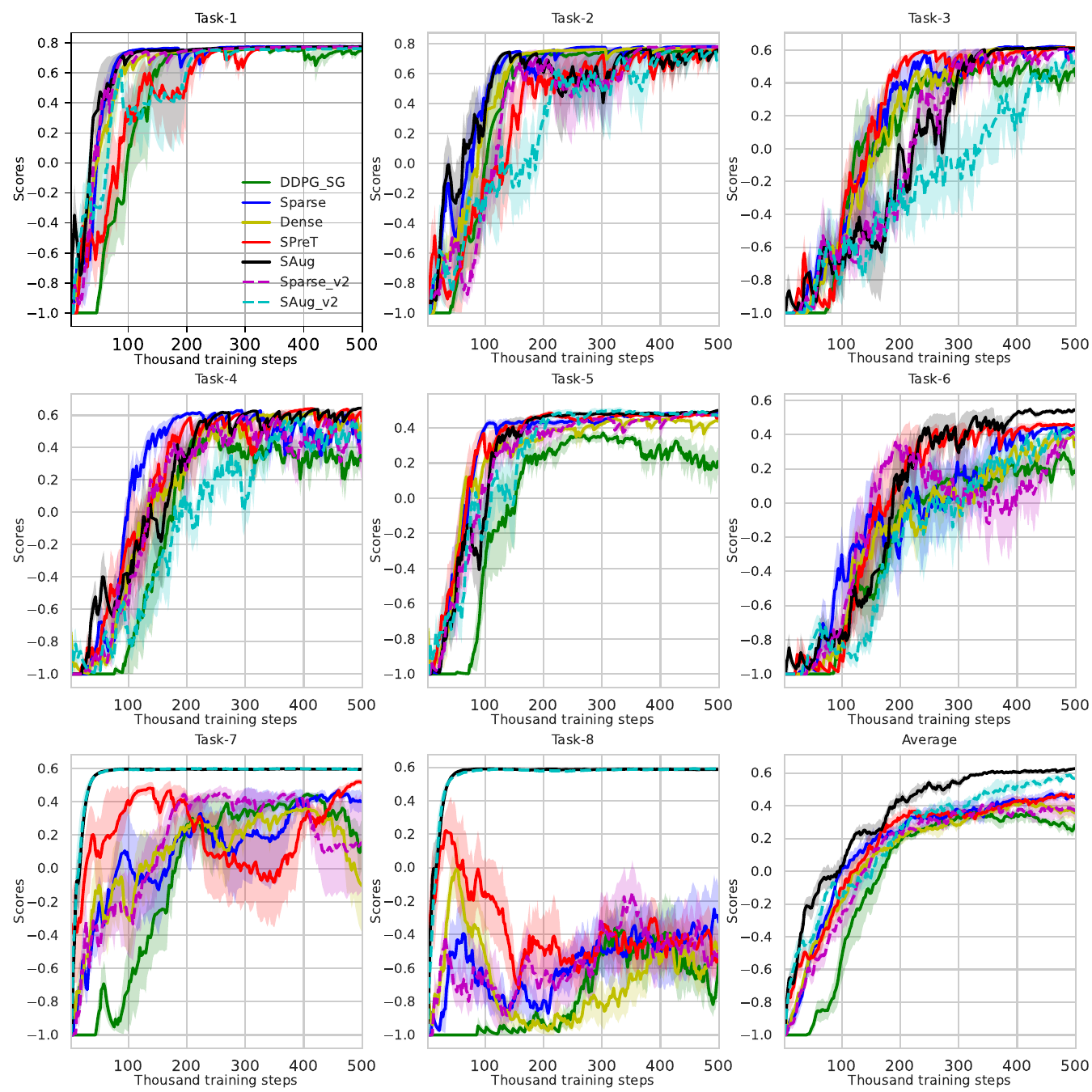}
	\caption{Learning curves of individual tasks (1-8) and average performance (9).}
	\label{fig5}
\end{figure}

The training process utilizes Stage\_ROS, a lightweight simulator designed for robot navigation. In the scenario augmentation process, the simulated robot's dimensions are scaled by $1/\rho$ relative to the real robot. This scaling ensures collision-free navigation in the simulated environment corresponds to successful traversal in the imagined scenario. Each episode begins with random robot placement in the unobstructed space of the $8\times8\text{m}^2$ training environment. The goal position is randomly selected from unobstructed areas within an $8\times8\text{m}^2$ window centered on the robot and Env1. Training efficiency is enhanced by implementing PID policy \cite{xie_learning_2018} for the initial 100 episodes, replacing random exploration. Following this period, the DRL agent executes stochastic actions sampled from the squashed Gaussian policy, initiating network updates. 
The evaluation procedure involves eight sequential navigation tasks, with start positions and goals indicated by blue and red circles, respectively, in Fig. \ref{fig4}. The performance metric score $S$, adopted from \cite{zhangIPAPRec}, is defined as:
\begin{equation}
\begin{aligned}
S =
\begin{cases}
1-\frac{2T_s}{T_{\text{max}}}, & \text{if success},\\
-1, & \text{otherwise}.
\end{cases}
\end{aligned}
\end{equation}
where $T_s$ represents the number of navigation steps required by the agent. This metric combines navigation time and task completion rate to quantify overall performance efficiency. Performance evaluation occurs at 2k-step intervals throughout the one-million-step training process. The training process is replicated five times with distinct random seeds to establish statistical significance. Fig. \ref{fig5} presents the learning curves with mean scores and variances, indicating that SAC\_SAug achieves higher performance metrics compared to SAC\_SAug\_v2.

\begin{figure*}[t]
	\centering\subfloat[REnv1]{
	\centering\includegraphics[width=0.21\linewidth]{figures/Figure7_1.pdf}\label{realEnv1}}
	\hfill\subfloat[REnv2-1]{
	\centering\includegraphics[width=0.21\linewidth]{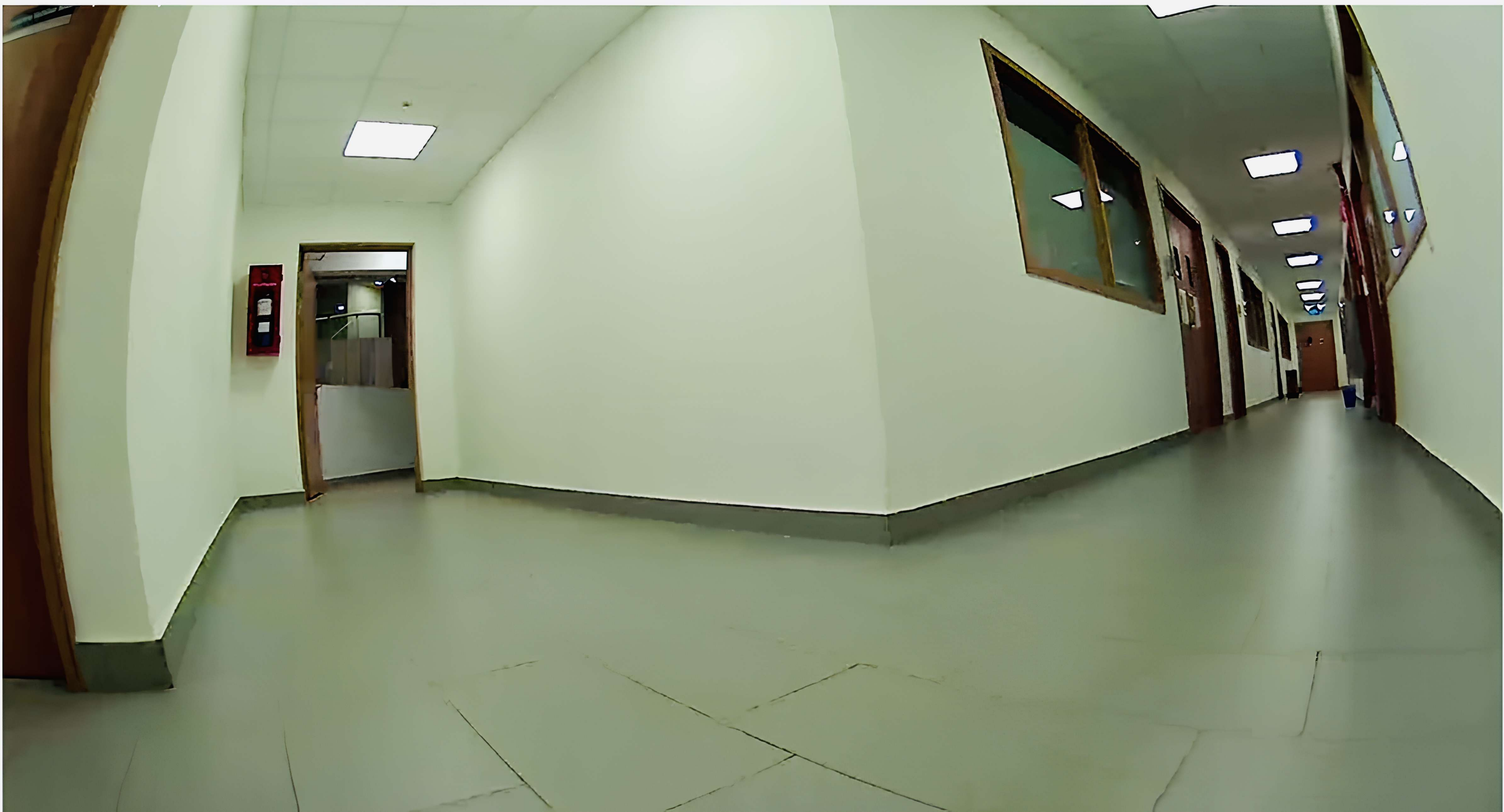}\label{realEnv2}}
	\subfloat[REnv2-2]{
	\centering\includegraphics[width=0.21\linewidth]{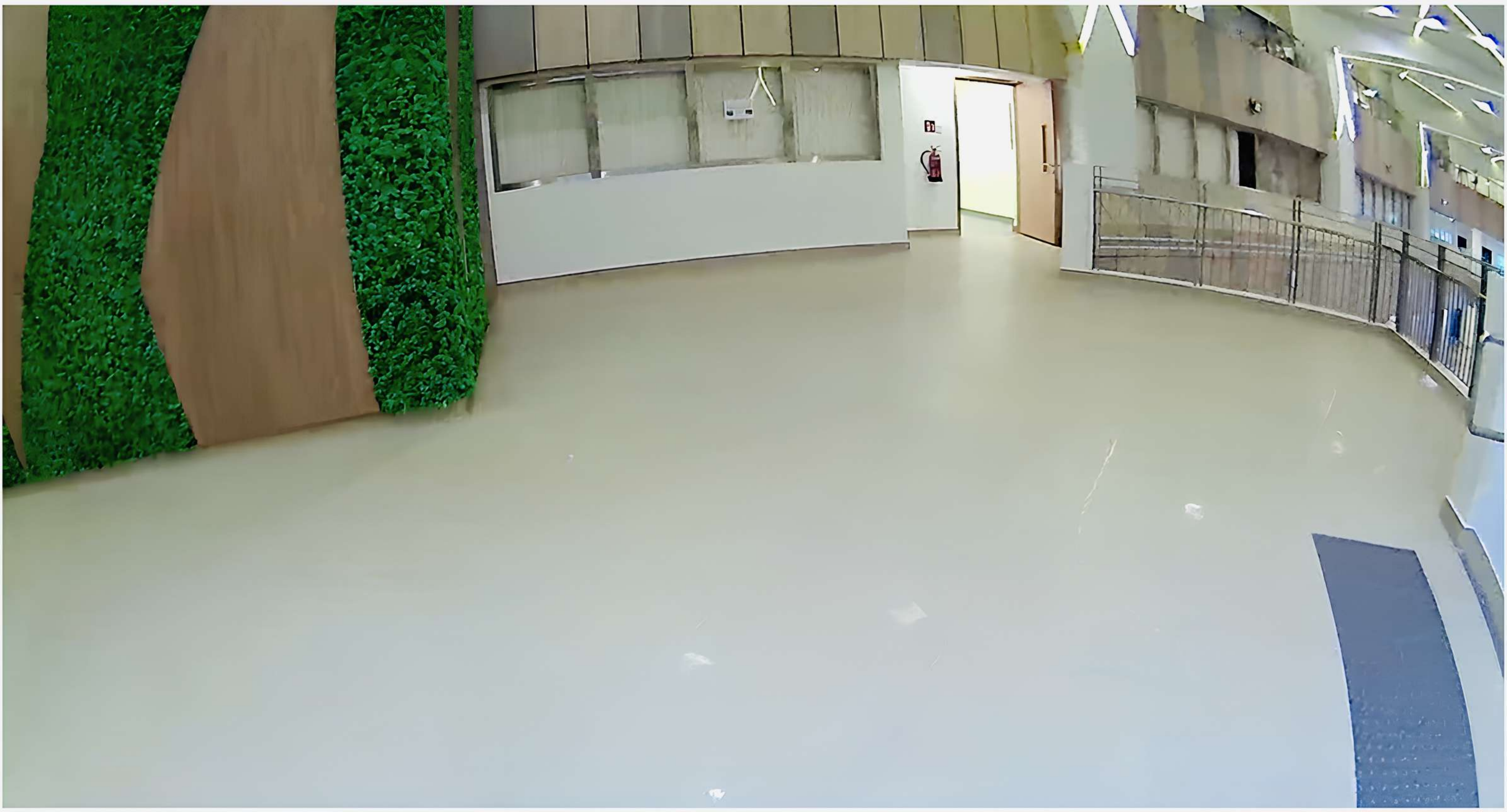}\label{realEnv3}}
	\hfill\subfloat[REnv3]{
	\centering\includegraphics[width=0.21\linewidth]{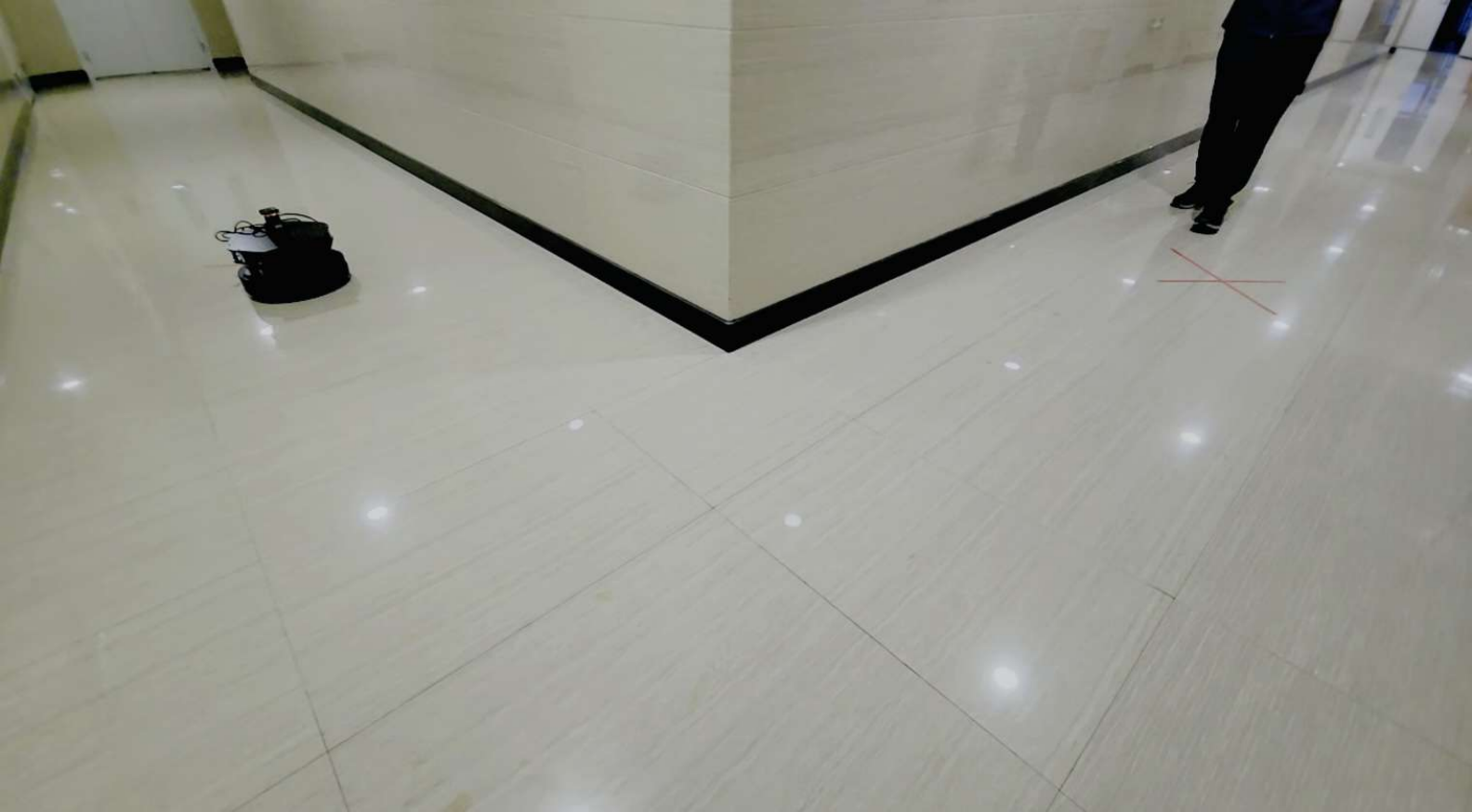}\label{realEnvped}}
    \hfill\subfloat[Test robot]{
	\centering\includegraphics[width=0.114\linewidth]{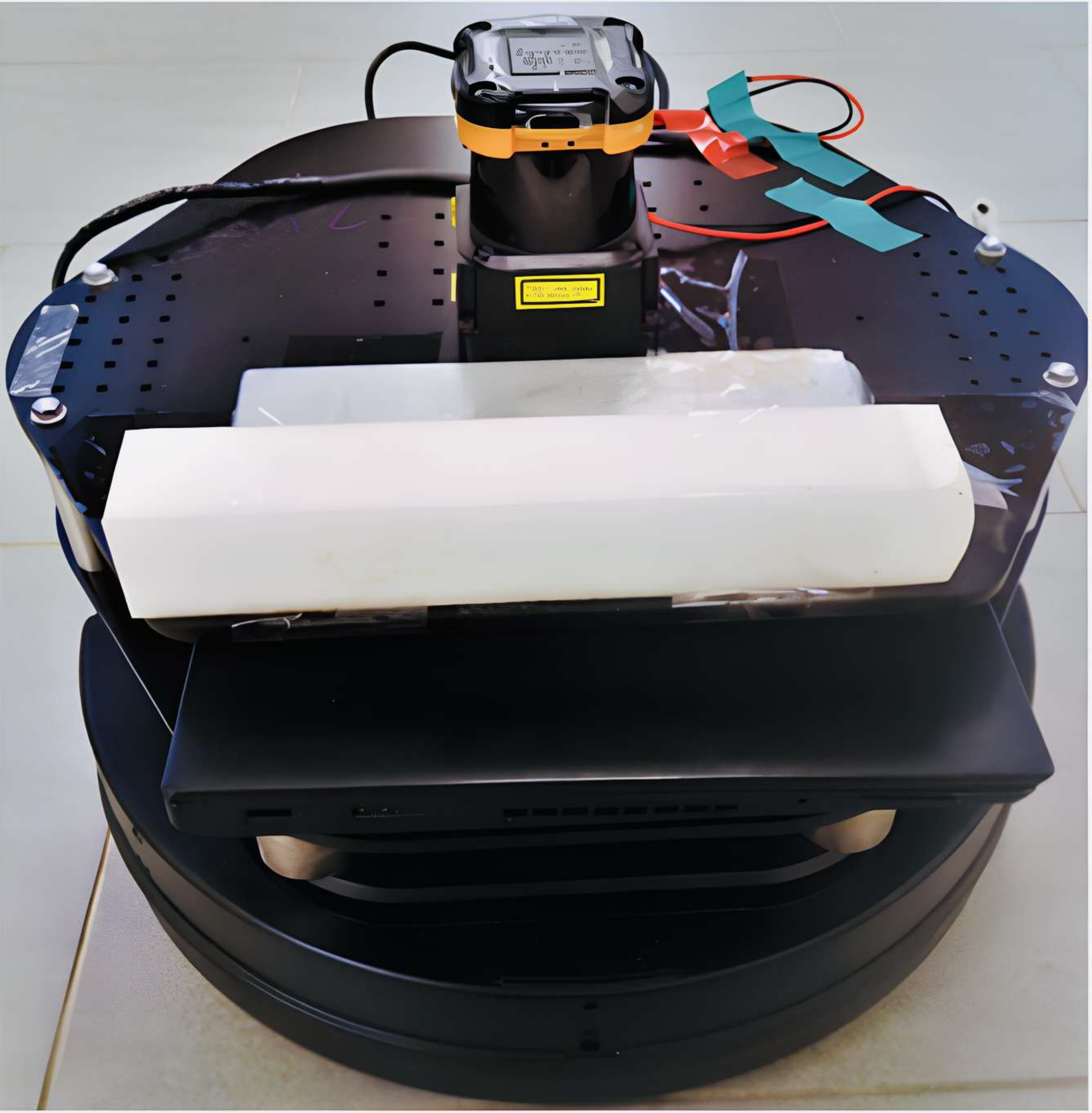}\label{robottur}}
	\caption{Real-world testing scenarios. (a) REnv1: Crowded indoor  scenario; (b) REnv2-1: Corridor scenario; (c) REnv2-2: Spacious indoor scenario; (d) REnv3: Corridor scenario with pedestrians. (e) The Turtlebot2 robot used for real-world testing. }
	\label{REnv}
\end{figure*}

The experimental results reveal significantly enhanced generalization performance through scenario augmentation, as demonstrated by consistent performance across all eight tasks. While all approaches achieved comparable performance scores in the training scenario (Env1, tasks 1 and 2), significant disparities emerged in novel environments. In tasks 7 and 8, approaches without scenario augmentation exhibited substantial performance instability and lower scores, indicating overfitting to the training scenario. Notably, SAC\_SAug maintained consistent high scores across all tasks, demonstrating superior generalization capability. The average performance curve shows SAC\_SAug achieving the highest final score compared to other methods.

SAC\_PreT-trained agents exhibited significant performance limitations, as shown in the learning curves. Despite pre-training in Env4, these agents demonstrated degraded performance in task 7 and task 8 , suggesting catastrophic forgetting. The learning curves also reveal that agents trained with dense rewards (SAC\_DR) exhibited more unstable learning patterns and inferior generalization compared to those using sparse rewards (SAC\_SR), particularly evident in tasks 5 through 8 where dense reward curves show higher variance and lower final performance. While both algorithmic choice and reward function design influence generalization, the comparative curves demonstrate that training across varied scenarios through SAC\_SAug constitutes the most effective approach for enhancing generalization performance.

\begin{figure*}[t]
    \centering
	  \subfloat[DDPG\_SG in REnv1]{
        \centering\includegraphics[width=0.1915
        \linewidth]{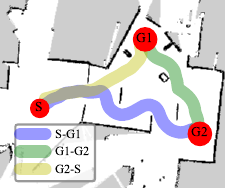}\label{env1ddpg}}
        \hfill
	  \subfloat[SAC\_SR in REnv1]{
        \centering\includegraphics[width=0.19\linewidth]{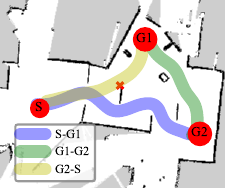}\label{env1sprse}}
        \hfill
        \subfloat[SAC\_DR in REnv1]{
        \centering\includegraphics[width=0.1912\linewidth]{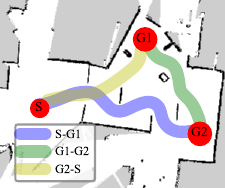}\label{env1dense}}
        \hfill
        \subfloat[SAC\_PreT in REnv1]{
        \centering\includegraphics[width=0.19\linewidth]{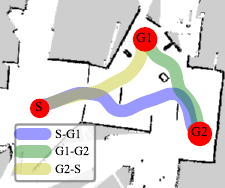}\label{env1PreT}}
        \hfill
        \subfloat[SAC\_SAug in REnv1]{
        \centering\includegraphics[width=0.1918\linewidth]{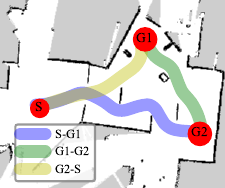}\label{env1SAug}}
        \\[-2ex]
        \subfloat[DDPG\_SG in REnv2]{
        \centering\includegraphics[width=0.1915
        \linewidth]{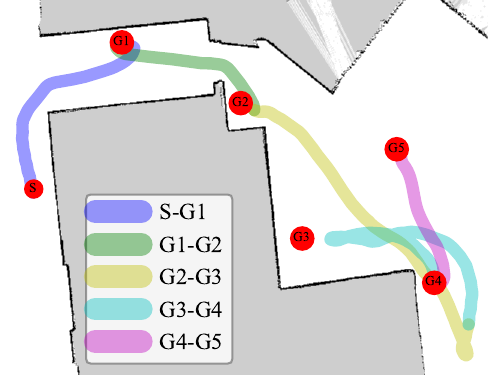}\label{env2ddpg}}
        \hfill
	  \subfloat[SAC\_SR in REnv2]{
        \centering\includegraphics[width=0.19\linewidth]{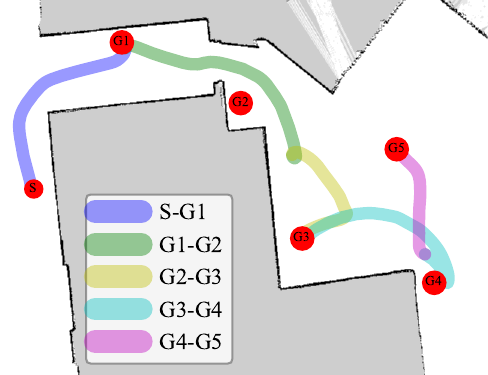}\label{env2sprse}}
        \hfill
        \subfloat[SAC\_DR in REnv2]{
        \centering\includegraphics[width=0.1912\linewidth]{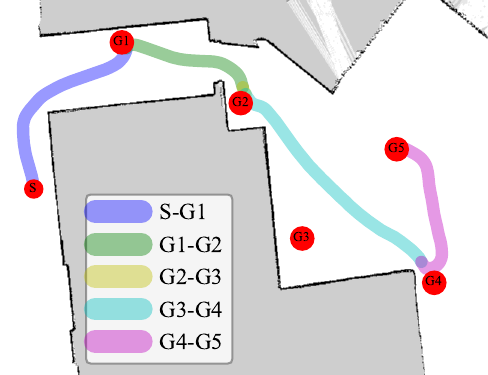}\label{env2dense}}
        \hfill
        \subfloat[SAC\_PreT in REnv2]{
        \centering\includegraphics[width=0.19\linewidth]{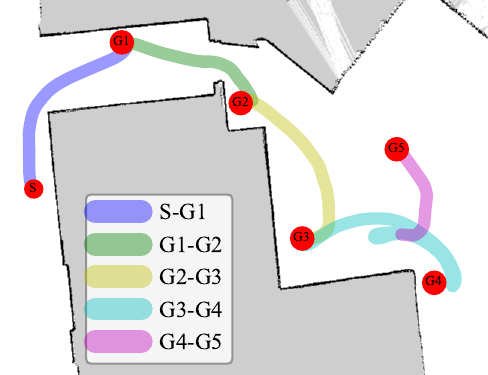}\label{env2PreT}}
        \hfill
        \subfloat[SAC\_SAug in REnv2]{
        \centering\includegraphics[width=0.1918\linewidth]{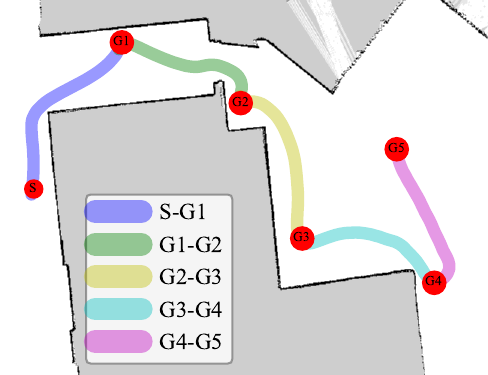}\label{env2SAug}}
        
	\caption{Robot trajectories for five approaches in REnv1 and REnv2. The experimental videos can be found in the supplementary
file.}
	\label{real_test}
\end{figure*}

\subsection{Navigation in Real-World environment}
Real-world environments present complex challenges and inherent uncertainties that computational simulations cannot fully replicate. To evaluate the performance of the five navigation approaches under real-world conditions, we conducted comprehensive experimental trials in physical environments. 

\subsubsection{Hardware Setup}
Experiments were performed using a Turtlebot2 robot equipped with a Hokuyo UTM-30LX LiDAR sensor, an onboard computer, and a pre-mapped environment for target localization, as shown in Fig. \ref{robottur}. The LiDAR sensor, with its wide $270^\circ$ FOV, 30-meter maximum range, and $0.25^\circ$ angular resolution, served as the primary perception device. Computational processing was executed on a cost-effective i7-7600U CPU laptop without GPU acceleration.
% The experiments were conducted using a Turtlebot2 robot equipped with a Hokuyo UTM-30LX LiDAR sensor, an onboard computer, and a pre-mapped environment for target localization, as shown in Fig. \ref{robottur}. The LiDAR sensor, with its wide $270^\circ$ FOV, 30-meter maximum range, and $0.25^\circ$ angular resolution, served as the primary perception device. The robot's onboard processing was handled by a cost-effective and accessible i7-7600U CPU laptop without a GPU.

Following an approach similar to \cite{tai_virtual--real_2017}, target localization was achieved using a pre-mapped testing environment, rather than dedicated sensors such as the UWB system employed in \cite{Fan2020DistributedScenarios}. The environment map was created using the \text{ROS GMapping} package, and robot localization within the map was achieved using \text{ROS AMCL}.  Subsequently, the map was utilized solely for localization purposes.

% The target's relative position was computed through coordinate comparison on the pre-built map, which was utilized exclusively for position determination and remained independent of motion planning, maintaining a distinct separation between localization and navigation functionalities.
% \begin{figure}[t]
%     \centering
%     \begin{tabular}{cc}
%         \subfloat[Turtlebot2 robot]{
%             \includegraphics[width=0.15\linewidth]{figures/robot11.pdf}\label{robottur}
%         } &
%         \subfloat[REnv1: a crowded indoor scenario]{
%             \includegraphics[width=0.238\linewidth]{figures/r1.pdf}\label{realEnv1}
%         } \\
%         \subfloat[REnv2-1: a corridor scenario]{
%             \includegraphics[width=0.238\linewidth]{figures/r2.pdf}\label{realEnv2}
%         } &
%         \subfloat[REnv2-2: a spacious indoor scenario]{
%             \includegraphics[width=0.238\linewidth]{figures/r3.pdf}\label{realEnv3}
%         }
%         \subfloat[REnv3: a pedestrian corridor scenario]{
%             \includegraphics[width=0.238\linewidth]{figures/outdr.pdf}\label{realEnvped}
%         }
        
%     \end{tabular}
%     \caption{Real-world testing robot and scenarios.}
%     \label{REnv}
% \end{figure}

\subsubsection{Testing scenarios and task description}
The real-world performance of the robot was assessed in three distinct environments, as depicted in Fig. \ref{REnv}. These environments, REnv1 (a crowded indoor space), REnv2 (a combination of a corridor section REnv2-1 and a spacious indoor area REnv2-2) and REnv3 (a corridor scenario with pedestrian traffic), were selected to evaluate the generalization capabilities of the DRL agents under varying levels of environmental complexity. In each environment, the robot was required to navigate from a starting point ``S'' to a series of target locations, denoted as $Gi$ and marked by red crosses on the ground (see Fig. \ref{REnv} and Fig. \ref{real_test}). 

% When the robot failed to reach a target, it was repositioned at the previous target point and reoriented towards the new target before initiating the next task.
% The real-world performance of the robot was evaluated in two distinct environments, as depicted in Fig. \ref{REnv}. These environments, REnv1 (a crowded indoor space) and REnv2 (a combination of a corridor section REnv2-1 and an open indoor area REnv2-2), were selected to assess the generalization capabilities of the DRL agents under varying levels of crowdedness. In each environment, the robot was tasked with navigating from a starting point ``S'' to a series of target locations, denoted as $Gi$ and indicated by red cross markers on the ground (see Fig. \ref{REnv} and Fig. \ref{real_test}). In cases where the robot was unable to reach a target, it would be repositioned at the previous target point and reoriented towards the new target before commencing the next task.

\subsubsection{Experimental Results}

\paragraph{Navigation Performance in Static Environment}

As shown, the robot trajectories and performance metrics are visualized in Figure \ref{real_test} and Table \ref{table1}, respectively. The metrics in the table are as follows: GR stands for the number of goals reached, TNN represents the total navigation time, and NC represents the number of collisions that occur during the robot's navigation process. In REnv1, SAC\_SAug demonstrated a navigation time of 29.4s, yielding a 14.3\% improvement over DDPG\_SG (34.3s).  All methods navigated to the three target points successfully, with the exception of SAC\_SR, which experienced one collision. Comparative analysis reveals that DDPG\_SG exhibited significant trajectory jerk, indicating control instability, while SAC\_SAug generated smoother paths with enhanced trajectory optimization and stable motion control.

\begin{table}[t]
\centering
\caption{Comparison of Navigation Metrics Across Five Approaches.}
\renewcommand\arraystretch{1.5}
\setlength{\tabcolsep}{1.2mm}
\begin{tabular}{cccccc}
\hline
\hline
Methods   & DDPG\_SG         & SAC\_SR         & SAC\_DR      & SAC\_PreT         & \textbf{\cellcolor{lightgray}SAC\_SAug}  \\ \hline
         & \multicolumn{5}{c}{REnv1}                                                               \\ \cline{2-6} 
\textbf{GR($\uparrow$)} & 3 & 3 & 3    & 3    &  \textbf{\cellcolor{lightgray}3}         \\
\textbf{TNN($\downarrow$)} & 34.3 & 30.6 & 30.9 & 32.1 & \textbf{\cellcolor{lightgray}29.4 }       \\ 
\textbf{NC($\downarrow$)}& 0  & 1 & 0   &  0   & \textbf{\cellcolor{lightgray}0} \\ \hline
         & \multicolumn{5}{c}{REnv2}                                                               \\ \cline{2-6} 
\textbf{GR($\uparrow$)} & 2 & 3 & 4 &  4  & \textbf{\cellcolor{lightgray}5}            \\
\textbf{TNN($\downarrow$)}   & 100.8 & 83.2 & 100.0  & 82.5 & \textbf{\cellcolor{lightgray}53.2}          \\
\textbf{NC($\downarrow$)}  & 0  & 0 & 0   &  0 & \textbf{\cellcolor{lightgray}0} \\
\hline
\hline
\label{table1}
\end{tabular}
\end{table}

In the more spacious environment REnv2, SAC\_SAug exhibited enhanced performance, successfully traversing all 5 target points with a navigation time of 53.2s, representing a 47.2\% reduction compared to DDPG\_SG (100.8s). Notably, in the challenging segments between G2-G3 and G3-G4, while other methods exhibited unnecessary detours and occasional failures in reaching target points, SAC\_SAug consistently generated optimal and smooth trajectories. As shown in Fig. \ref{env1SAug} and Fig. \ref{env2SAug}, SAC\_SAug demonstrated superior performance across both environments, featuring near-optimal path planning, smooth motion profiles, and robust environmental adaptability.

\begin{figure}[t]
	\centering\subfloat[Single Pedestrian Co-Directional]{
	\centering\includegraphics[width=0.48\linewidth]{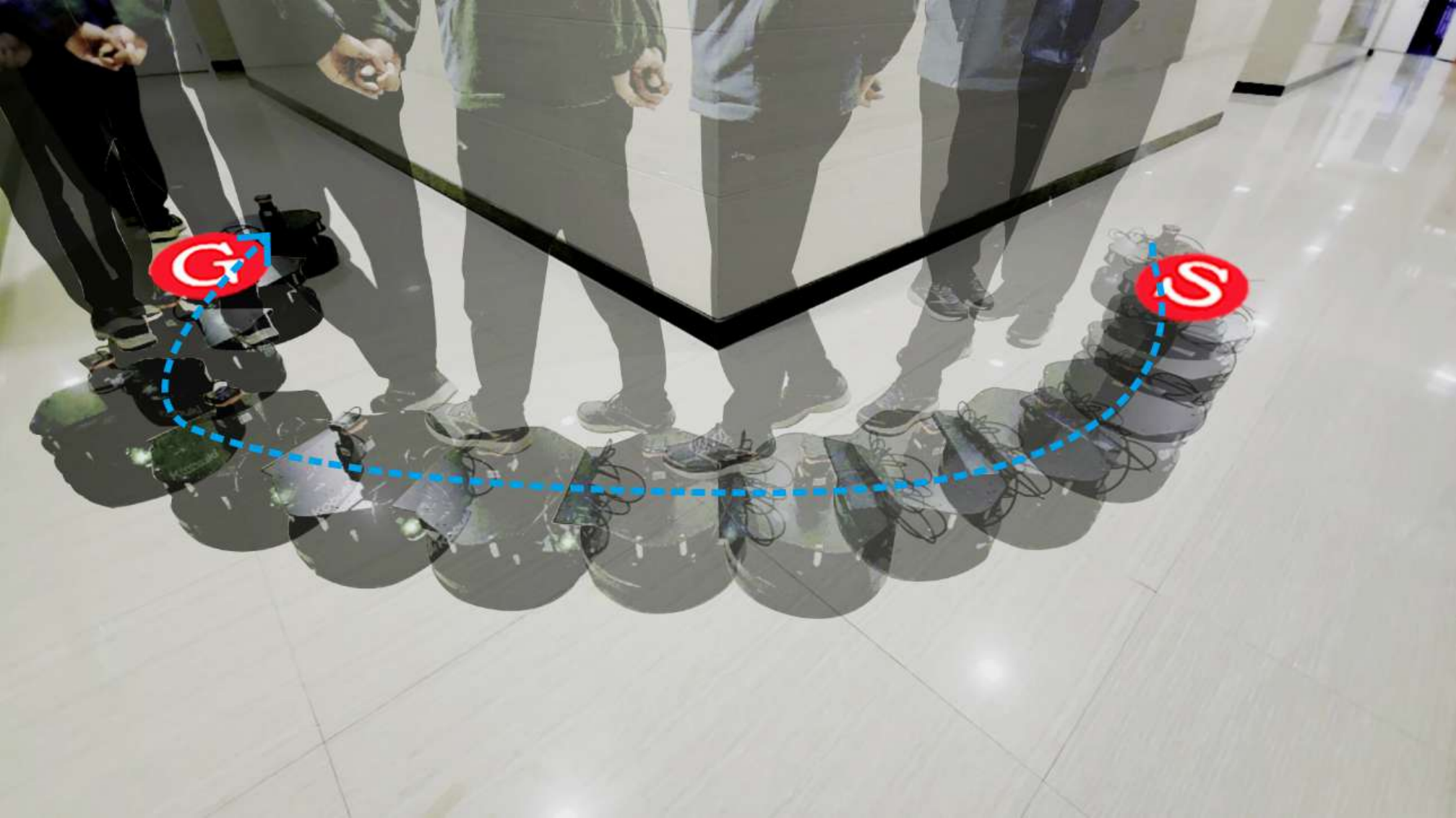}\label{danren1}}
	\hfill\subfloat[Sudden Pedestrian Obstruction]{
	\centering\includegraphics[width=0.48\linewidth]{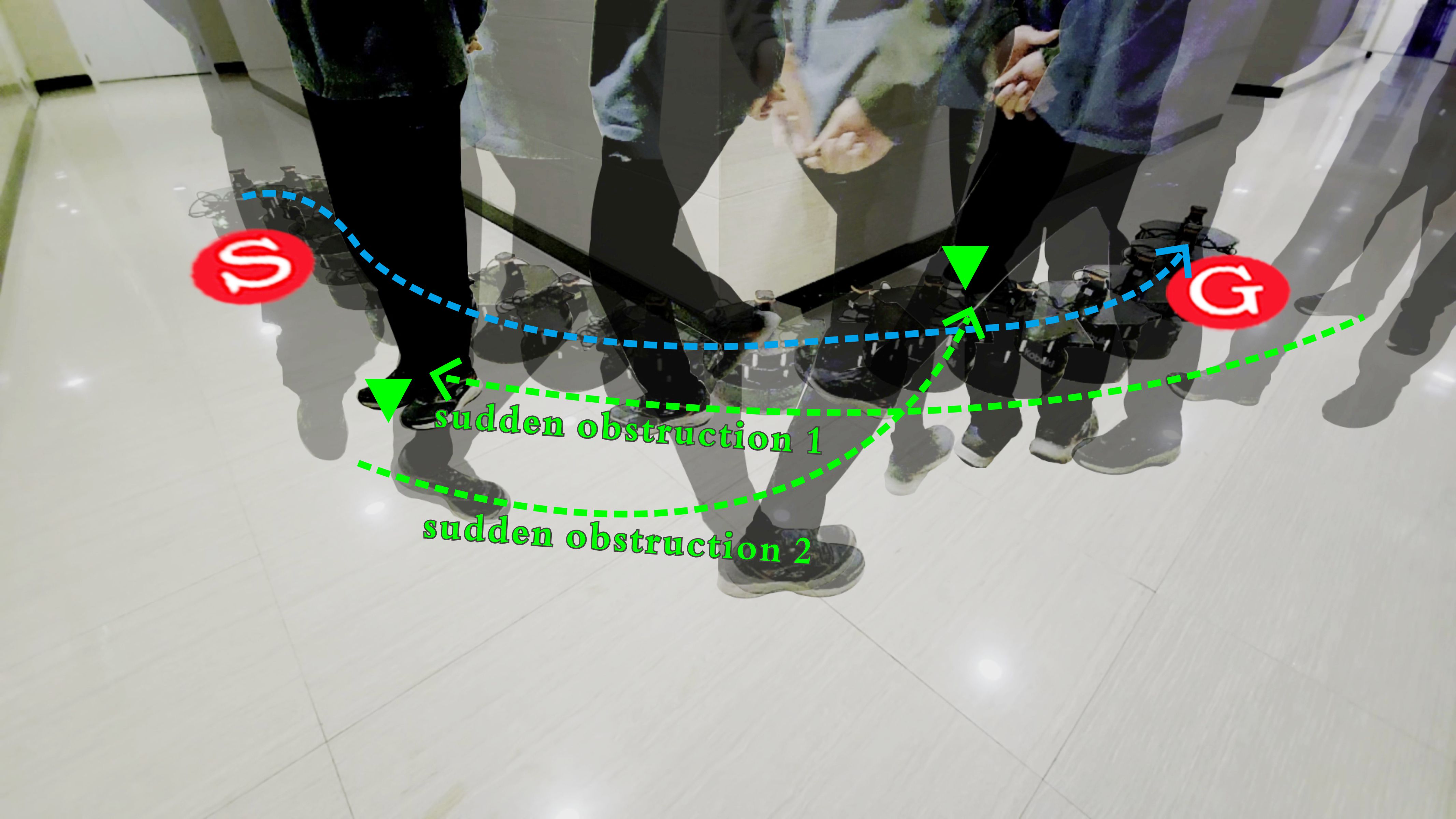}\label{danren2}}
	\\[-0ex]
	\subfloat[Two Pedestrians Oncoming]{
	\centering\includegraphics[width=0.48\linewidth]{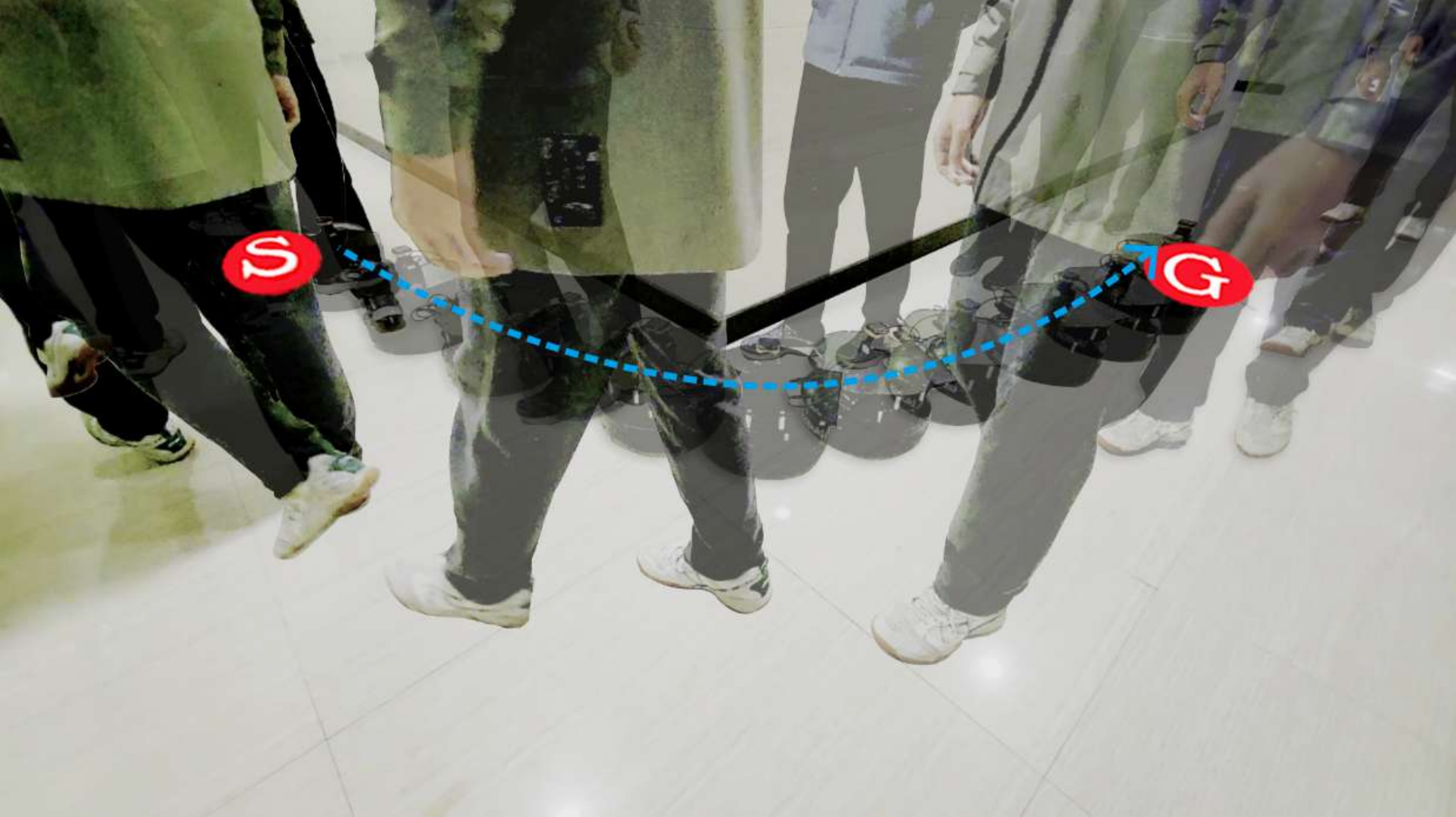}\label{shaungren1}}
	\hfill\subfloat[Two Pedestrians Bidirectional]{
	\centering\includegraphics[width=0.48\linewidth]{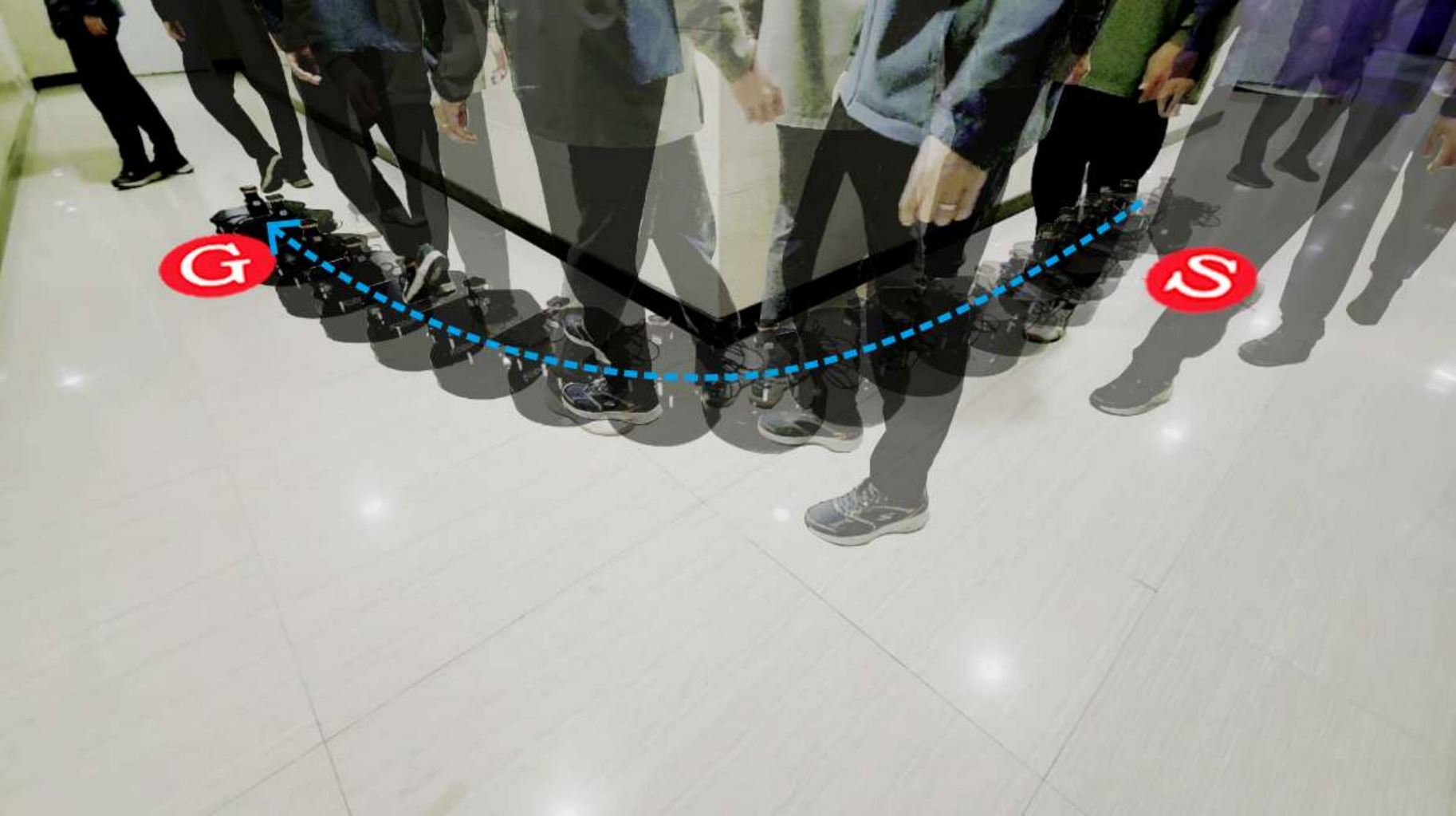}\label{shuangren2}}
	\caption{Real-world robot trajectories in REnv3. The experimental videos can be found in the supplementary
file.}
	\label{real_test}
\end{figure}

\paragraph{Navigation Performance in Dynamic Scenarios}

To evaluate the method’s capability in handling dynamic obstacles, we established four distinct dynamic pedestrian test scenarios: co-directional movement with a single pedestrian, sudden obstruction, counter-directional interaction with two pedestrians, and navigation through opposing pedestrian flows. As shown in Fig. \ref{real_test}, SAC\_SAug exhibited effective performance across all scenarios: maintaining optimal following distances and velocity adaptation in co-directional movement, executing efficient evasive maneuvers in obstruction cases, implementing safe navigation between opposing pedestrians, and achieving stable trajectories through pedestrian flows. 
The experimental results validate the key capabilities of SAC\_SAug: adaptive response to dynamic obstacles, efficient goal-reaching behavior, and consistent performance across varied environmental conditions. 
% The results validate SAC\_SAug's effectiveness in dynamic environments, demonstrating applications in dynamic environments.

% \begin{figure*}[t]
%     \centering
%     \begin{tabular}{cc}

%         \begin{tabular}{cc}
%             \subfloat[Single Pedestrian Co-Directional Scenario]{
%                 \includegraphics[width=0.33\linewidth]{figures/danren_tongxaing.pdf}\label{danren1}
%             } &
%             \subfloat[Sudden Pedestrian Obstruction Scenario]{
%                 \includegraphics[width=0.33\linewidth]{figures/danrenzhe.jpg}\label{danren2}
%             } \\
%             \subfloat[Two Pedestrians Oncoming Scenario]{
%                 \includegraphics[width=0.33\linewidth]{figures/shuangren_tongxaing.pdf}\label{shaungren1}
%             } &
%             \subfloat[Two Pedestrians Cross-Directional Scenario]{
%                 \includegraphics[width=0.33\linewidth]{figures/shuangren_xiangxaing.pdf}\label{shuangren2}
%             }
%         \end{tabular}
%     \end{tabular}
%     \caption{Real-world robot trajectories in REnv3.}
%     \label{pedREnv}
% \end{figure*}

\section{Conclusion}
In this paper, we introduce scenario augmentation, a novel method to enhance the generalization performance of DRL navigation agents. By mapping observations into an imagined space and remapping actions, this technique facilitates diverse training without altering the actual environment. Through extensive experiments in both simulated and real-world environments, we demonstrated that insufficient diversity of training scenarios is the primary factor limiting navigation performance. Our comparative experiments clearly showed that scenario augmentation significantly improves the agent's adaptability to unfamiliar environments.
The real-world experiments further validated the practical value of our approach, with the scenario augmentation method achieving both higher success rates and shorter navigation times compared to baseline approaches. Moreover, it showed excellent obstacle avoidance capabilities and navigation performance in complex dynamic environments.

% \section*{Acknowledgments}
% This paper was supported by Ningbo ``Science and Innovation Yongjiang 2035'' Key Technology Breakthrough Program (Grant No. 2024Z127). The manuscript was polished with the assistance of AI language tools.


\bibliographystyle{IEEEtran}
\bibliography{mylib}

\end{document}